# Half-CNN: A General Framework for Whole-Image Regression

Jun Yuan
National University of Singapore
Singapore 117583
yuanjun@nus.edu.sg

Bingbing Ni
Advanced Digital Sciences Center
Singapore 138632
bingbing.ni@adsc.com.sg

Ashraf A.Kassim
National University of Singapore
Singapore 117583
ashraf@nus.edu.sg

## Abstract

*The Convolutional Neural Network (CNN) has achieved great success in image classification. The classification model can also be utilized at image or patch level for many other applications, such as object detection and segmentation. In this paper, we propose a whole-image CNN regression model, by removing the full connection layer and training the network with continuous feature maps. This is a generic regression framework that fits many applications. We demonstrate this method through two tasks: simultaneous face detection & segmentation, and scene saliency prediction. The result is comparable with other models in the respective fields, using only a small scale network. Since the regression model is trained on corresponding image / feature map pairs, there are no requirements on uniform input size as opposed to the classification model. Our framework avoids classifier design, a process that may introduce too much manual intervention in model development. Yet, it is highly correlated to the classification network and offers some in-deep review of CNN structures.*

## 1. Introduction

The image classification techniques have evolved vastly in recent years, from Bag-of-Visual-Words (BoVW) [1, 2], Fisher Vector (FV) Encoding [3], to the state-of-the-art Convolutional Neural Network (CNN) [4]. The BoVW and FV models typically consist of a pipeline, with cascading processes of feature description, dictionary building and feature encoding, pooling over image and classification. A linear SVM classifier is used at final stage, and the whole process can be considered as a non-linear classification model on images. These techniques have been extensively studied in [5, 6]. The CNN classification model consists of several convolution blocks followed by fully connected layers. Each convolution block consists of a convolution layer with a non-linear activation, a pooling layer, and sometimes a local normalization layer [7]. The fully connected layer can be considered as a final pooling layer, which feeds the image feature to the classifier. The CNN model typically uses a softmax classifier at the training stage. After training, the softmax classifier is often replaced by an SVM classifier on the final image representations, for improved classification performance. Sometimes the SVM is directly used in the network training process.

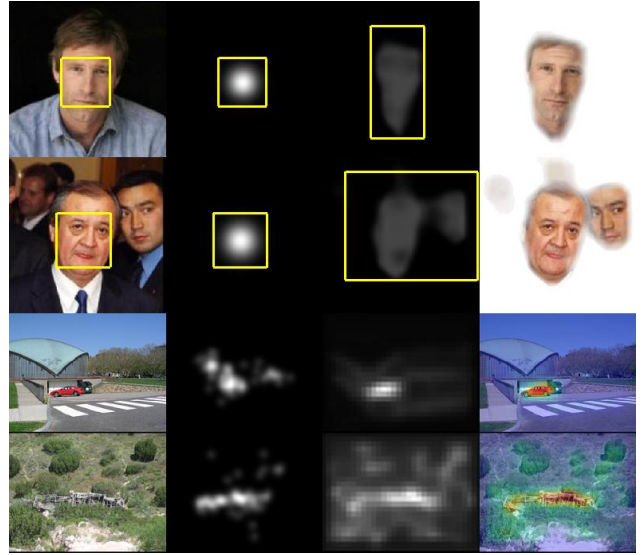

Figure 1: Example applications of the CNN regression model. The top two rows demonstrate our model on simultaneous face detection and segmentation, and the bottom two rows demonstrate our model on scene saliency prediction. The columns correspond to the input images, ground truth feature maps, network output and results interpretation.

The key advance of CNN lies in its ability to learn adaptive features, as opposed to the hand-crafted features like SIFT [8] in BoVW and FV models. These hand-crafted features are intuitive for human perception; they can also be considered as a shallow network. The former convolution blocks in CNN perform similar tasks as these feature detectors; however the learned features are more optimized for the given task. Similar statements hold for the encoding process. The linear SVM classifier at output is also a shallow network. Thus the classification pipeline can be treated as a (relatively) shallow network consisting of hand-crafted function blocks, whereas CNN is a deep network consisting of optimized blocks. This explains the great success of CNN in classification tasks. The CNN classification model achieves state-of-the-art performance on classification datasets such as ImageNet ILSVRC [9] and PASCAL VOC challenges [10]. A comprehensive study can be found in [11]. The pre-trained ImageNet model [12, 13, 14] is also successful in generalizing to other image classification datasets.

The CNN is also widely applied on non-classification tasks, such as object detection [15, 16] and segmentation [17, 16]. The model can be either applied on image level, or patch level with information aggregation. The ImageNet pre-trained model is also widely used in these applications.

However, it seems that CNN is not so widely used in regression tasks, which is also a fundamental problem in machine learning. Regression and classification problems are highly correlated [18], and can be transferred to the other in many scenarios. For example, the softmax classifier is related to the softmax regression [19], and the SVM classifier is based on geometric regression. It is natural to develop a deep CNN model for regression tasks as a counterpart of classification.

In this paper, we propose a simple but generic 2D CNN regression model. By removing the fully connected layer(s) in the classification network, we yield a network output that is locally correlated with network input (due to the locality of convolution operation); this is also the key characteristic of local regression models. The final output feature map is generated through a linear combination of convolution channels, which can be considered as a partially connected layer. Thus we can get a 2D local regression network out of the classification network. We also propose an up-sampling layer to reduce the down-sampling effects from pooling operations. The relationship between the classification and regression networks is discussed in detail in section 5.

The ground truth to train the regression network are feature maps generated from input images. The size of the maps is related with input images, either identical or with a pre-defined down-sampling factor. These two scenarios correspond to the convolution and pooling operations in the neural network. Since there is no fully connected layer, our model does not pose a requirement on identical input size, which can be a problem in the CNN classification model.

We apply the above framework to two applications: simultaneous face detection & segmentation and saliency prediction. In the former task, the ground truth feature maps are generated by fitting a 2D Gaussian density function inside the detection window. The trained network can not only recover the position of faces, but also output the segmented face regions on the feature map. The saliency prediction is a natural application of our network, since the ground truth is already given in real-value feature maps. We demonstrate the effectiveness of our work through a small scale regression network.

## 2. Related Work

Face detection is an important task in computer vision. There are various models from the appearance to learning based models [20, 21, 22, 23]. The classical models are extensively studied in [24]. The CNN models are recently developed into this field [25]. This paper uses convolution to generate feature maps, and a SVM classifier to detect face windows. This approach is similar to the part-based model object detection [26]. Some other face detection algorithms are implemented through segmentation [27, 28]; the idea is close to our framework. However, our network does not need to include any specific knowledge for facial features, e.g. skin color, to the detection or segmentation task. The features are fully learned without manual crafting in our network.

Face detection can also be combined with simultaneous face landmark localization [29, 30, 31]. Part of our work is based on Y. Sun et al. [31], where a CNN model is applied to detect face landmarks. A three-level cascaded network is used to predict face landmark through a coarse-to-fine localization process. Their network also uses a regression model, but different from ours. In their approach, the network outputs several locations as a fixed-size set of real numbers, which can be considered as a mapping from the 2D real domain to the 1D real domain, i.e. a 2D-1D mapping. The CNN classification model maps images to a fixed finite countable set which can be considered as a 2D-0D mapping. A fully connected layer is needed whenever there is dimension reduction from the input to the output. As a comparison, our CNN regression model is a 2D-2D mapping between images and feature maps, where the full connection is avoided. The work [31] can be considered as a bridge between the classification CNN model and our model.

We also test the face landmark localization in our framework. However, this task heavily depends on the knowledge of spatial relationship between face landmark points, and a pure texture approach like ours to give independent predictions of landmarks is not ideal. The CNN localization model in Y. Sun et al. [31] also utilizes the spatial relationship of landmarks implicitly through a global-to-local refinement. We shall briefly discuss this point in section 4.

Saliency prediction is another important study field, serving as a bridge between computer vision and human perception. The saliency prediction models evolved from Itti [32], GBVS [33], detection / segmentation models [34, 35], and to more recent CNN-based models [36, 37]. An extensive study of classical models can be found in [38]. A comparison of these models can be found in [39].

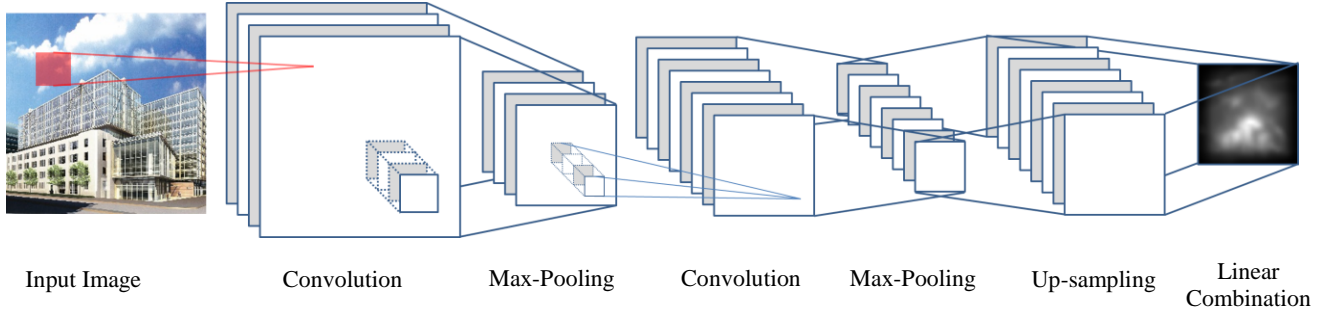

| Input Image | Convolution | Max-Pooling | Convolution | Max-Pooling | Up-sampling | Linear Combination |

Figure 2: A Schematic of CNN Regression Model. The up-sampling layer is used to maintain the feature size, and the output is produced by a linear combination on convolution channels. These two layers are specific in our framework.

There are two types of saliency models: bottom-up and up-to-down saliency model. The former is typically more focused on human perception process and low level image information, whereas the latter is more related to scene knowledge on interest objects. The latter is also highly related to object detection and segmentation; some models explicitly use detection and segmentation techniques to give saliency predictions [34, 35]. The recent CNN models also offer competitive performance in this field. The eDN [37] uses a large number of randomly initialized CNNs, picking those with good performance and aggregates their output feature maps to give predictions. The latter Deep Gaze I [36] uses ImageNet pre-trained network without the full connection layer, and learns a weight on linear combination of convolution channels. This model offers the state-of-the-art performance. Our model has a structure similar to the Deep Gaze model, but trained in a different way without using classification information.

Since saliency information is provided as feature maps, this is a natural application fits into our framework. We use the CNN regression model to train the saliency model and give comparative studies in section 4. There is also a detailed discussion in section 5.

## 3. Implementation Details

Our network consists of several convolution blocks and an output layer. The convolution blocks follow structure of [7], and the output layer combines all the feature channels through a linear combination.

**Convolution blocks**. Each convolution block includes a convolution layer, a ReLU activation, and a pooling layer with a down-sampling factor of 2. Max-pooling is used in as the pooling function. A normalization layer introduced in [7] can also be used after the pooling layer, with the local normalization function below:

$$b_{x,y} = a_{x,y} / \left( k + \alpha \sum_{j \in N(j)} (a_{x,y}^2) \right)^\beta$$

The above function computes the feature response at location $(x, y)$ by normalizing the responses at the same location in its $N$ neighboring channels. In practice, the normalization layer can offer limited performance gain, and is used only in the first two convolution blocks in our experiments.

There is a limit on the amount of pooling layers that can be used in our network, due to the size limitation of the ground truth feature maps. For example, if the feature map is 4× times smaller than the original image, we can only use two max-pooling layers with a down-sampling factor 2. There are several ways to get around this: We can either to use pooling with a stride of 1, or use convolution without pooling in the following blocks. However, these two methods reduce the non-linearity of the network, which is not a desired property.

In this paper, we propose an up-sampling layer, to maintain the size of intermediate features in the network. The up-sampling layer follows the pooling layer and restores the size of pooled features, while retaining the non-linearity of the network. The up-sampling function is defined as follows:

$$A_{i+1}(px - p + 1:px, py - p + 1:py) = A_i(x, y)$$

where $p$ is the down-sampling factor used in the pooling layer, typically of the value 2. The above function copies a value from the pooled features to a $p \times p$ block in the following up-sampling layer. The back-propagation rule of this layer is just the average-pooling layer in the reverse direction, with a scale of $p^2$. Since each block in the up-sampling layer consists of the same value, the back-propagation principal of this layer can be simplified as:

$$dA_i(x, y) = p^2 \, dA_{i+1}(px, py)$$

**Output Layer**. The output layer combines the feature channels from the last convolution block through a linear combination. Then a sigmoid activation function is applied to produce the final output. The layer forward and back propagation rules are shown below:

$$A_o = \text{sigm}\left(\sum_i w_i A_i + b\right) \triangleq \text{sigm}(Z)$$
$$dZ = (M \circ dA_o) \circ A_o \circ (1 - A_o)$$
$$dw_i = \sum_{x,y} A_{i,x,y} dZ_{x,y}, \quad db = \sum_{x,y} dZ_{x,y}$$
$$dA_i = w_i dZ$$

where $A_i$ is the convoluted feature channels, $w$ and $b$ are the weights and bias of the linear combination. $M$ is a mask indicating the content in images and feature maps, which is introduced in section 3.1.

Also, it is possible to use other type of combinations, such as a max-pooling function among all the convolution channels. However, this function seems to underperform the linear combination in our experiments.

Since the feature maps consist of real continuous values (typically ranging from 0 to 1), the network performs a 2D-2D regression task. The structure of our network is shown in Figure 2.

### 3.1. Size considerations in practice

The size of ground truth feature map is typically related to the input image. Because there are no fully connected layers in the regression framework, we can bypass the uniform input size requirement in typical classification CNNs. If the input sizes are different in the classification model, the images have to be warped or cropped to the same size. Different image aspect ratio is also a problem. However, our regression model can fundamentally avoid these headaches.

The convolution operation does not depend on image size; however the layer input needs to be properly padded according to the filter size, to ensure the correct down-sampling factor after pooling. The input and output size of convolution blocks differ by a factor of 2, if pooling without up-sampling is used. When an up-sampling layer is present, the input and output sizes of convolution blocks are identical.

In theory, the network can accept input images of all different sizes. The batch gradient can be calculated by averaging all the individual gradients produced by single images in the batch. However, for fast computation, it is recommended that all the images within a batch are identical in size to utilize the advanced data structure in typical CNN implementations.

Should the input sizes differ vastly, there is another convenient way to bypass the uniform-size requirement. All the input images in a batch can be padded to the same size, with a mask recording the content of each input. The ground truth feature maps are padded in accordance with corresponding images. The mask is then applied to the derivative of the output feature map to ensure correct gradient computation. This technique cannot be applied on the classification network, since the fully connected layer requires proper arrangement of features. Applying masks will interrupt the pre-defined feature sequences.

The above padding-masking scheme can also be applied on filters, if a convolution layer includes filters of different sizes. However, this seems to be a relatively rare setup in most of the CNN applications.

## 4. Experiments

We use the above framework for two applications. One is simultaneous face detection & segmentation on the LFW face dataset [40], the other is saliency prediction on the MIT dataset (1003 images version) [41]. The two networks are both trained on the MatConvNet platform [42], a Matlab toolbox based on the famous CAFFE [13] CNN implementation.

### 4.1. Face Detection & Segmentation

We use 5590 face images from the LFW dataset, with different viewing angles. All images are of identical size 250×250. We use 4151 images for training and the rest for testing. The detection window provided in [31] is used as the ground truth.

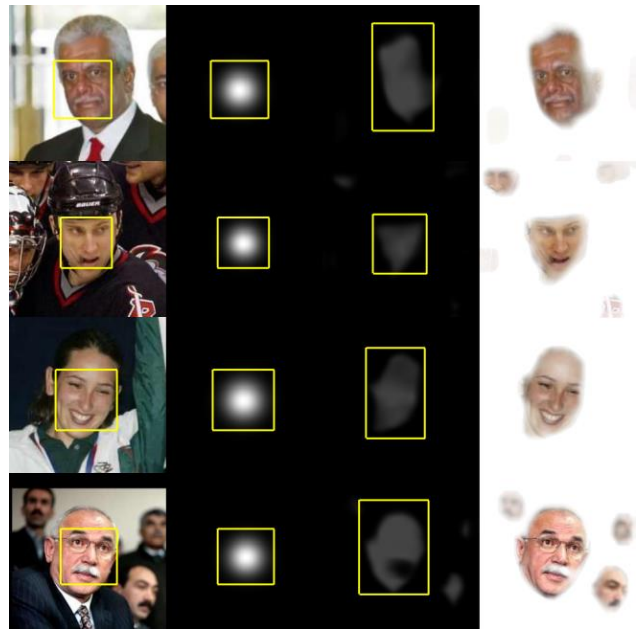

Figure 3: Examples of the CNN regression model on simultaneous face detection & segmentation experiment. The columns correspond to the input, ground truth, output features and results interpretation. The network is able to detect and segment multiple faces.

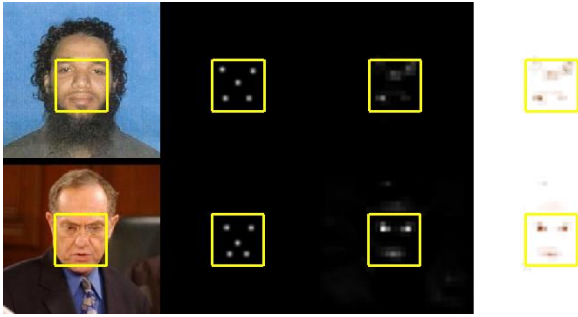

Figure 4: Examples of the CNN regression model on face landmark localization. The network does not perform well because the spatial relationship between facial landmarks is not taken into consideration.

We use a simple technique to generate the ground truth feature maps: a 2D Gaussian density function is fit at the center of the detection window, with the diameter ($6\sigma_x$ and $6\sigma_y$) being the window width and height. The size of input images is padded to 256×256 as stated in section 3, and a 4× down-sampling is applied to the generated feature maps.

We use 3 convolution blocks in this experiment. The first two blocks consist of 5 filters of size 11×11 and 7×7 respectively, with max-pooling and normalization. The following block consists of 5 filters of size 5×5, with up-sampling layer after max-pooling.

For numerical stability, the ground truth feature maps are re-normalized to [0.1, 0.9] for the sigmoid activation at the output layer. Also, there is a small $L_2$ penalty on filter weights and biases of each layer for regularization. This penalty terms can prevent the network from over-fitting and improves generalization ability.

Some sample outputs are shown in Figure 3. The output feature map is overlayed on input images for visualization. It can be seen that the output feature maps correspond to the segmentation of faces in the input. The neck area is sometimes included in the feature map; this is natural because of the color similarity between face and neck areas. Increase the network complexity might relieve this problem.

The detection window can be retrieved by analyzing the feature maps: get the center and standard deviation of the response, and fit the window as the reverse process of generating Gaussian density functions. The retrieval rate of detection window is greater than 95% in our experiments.

Though we train the network with a simple Gaussian density function in the detection window, the network surprisingly provides the ability of both detection and segmentation, with only a small scale network. Moreover, the network can detect and segment multiple faces as shown in Figure 1 and 3. This indicates the generalization ability of our regression network.

We also tested our regression model on face landmark localization experiments, based on the work of [31]. The ground truth feature maps are generated by fitting small 2D Gaussian density functions to the face landmark positions. However, our network does not perform well on this application, for that it is a more textural approach, without incorporating any spatial relationship between landmarks. This relationship is crucial in this application, and detecting the landmarks independently through our framework is not ideal. Some sample detection results are shown in Figure 4.

The landmark localization network is able to detect more structural regions like eyes, and fails in the less structural regions on noses. This is expected because of the color similarity in nose regions. The structural contents in eyes are most, and moderate in mouths; noses offer the least information in textures.

### 4.2. Saliency Prediction

The MIT saliency dataset consists of 1003 indoor and outdoor scene images. The longest dimension of each image is 1024 pixels, and the other edge ranges between 405 and 1024 pixels. The aspect ratio is typically around 4:3. The ground truth of saliency prediction is already provided as feature maps, thus this is an application that naturally fits our regression framework. We pad the images to 256×256, and use a down-sampling factor of 4 for feature maps. We use the same network regularization parameters as in the former face detection & segmentation experiments.

We use 4 convolution blocks in this experiment. The first two blocks consist of 10 filters of size 7×7, with max-pooling and local normalization. The following two blocks

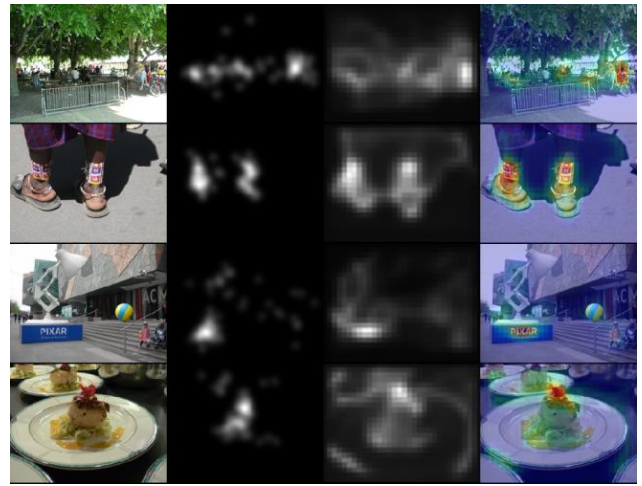

Figure 5: Examples of the CNN regression model on scene saliency prediction on MIT 1003 dataset.

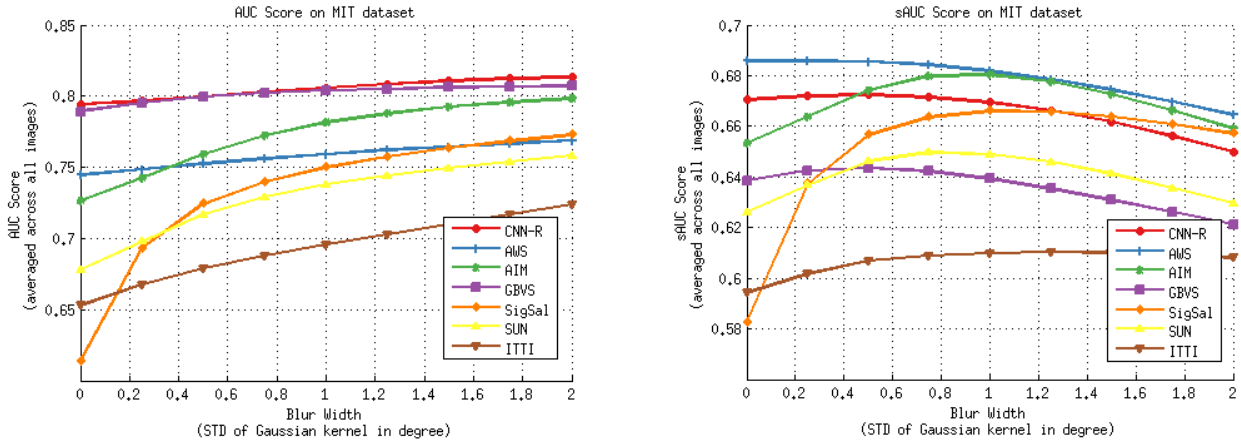

Figure 6: Comparison with classical saliency models, under the AUC and sAUC performance metric.

consist of 10 filters of size 5×5, with max-pooling and up-sampling.

Some experiment results are shown in Figure 5. We can see that the network output is highly correlated with the ground truth saliency map. Some quantitative evaluation and comparison with other methods are shown in Figure 6.

Our model outperforms most of the classical saliency prediction methods, while not incorporating any specific knowledge to saliency studies. We have yet to compare our model to the latest work, due to the difference in database and evaluation methods.

Additional experiment results can be found at Figure 7 and 8 for both applications.

**Computation Time**. Since our regression network is small in scale, the L-BFGS [43] algorithm is used for fast convergence. The network can be trained in 2~3 hours and 5~6 hours respectively for the face detection & segmentation and saliency prediction, with an NVidia K40 graphics card.

## 5. Discussion

**Generality**. Our CNN regression model is a general framework that can be applied on a variety of applications, where the output is a feature map correlated with input images. For example, other than our simple face detection & segmentation, we can generate feature maps on more complicated segmentation tasks such as the PASCAL VOC challenge [10] and neuron membrane segmentation [44] in medical images, with the segmented object as the ground truth. This is part of our future study on more advanced detection & segmentation applications.

Also, our framework provides another view of classical features, such as SIFT and HoG [45]. The output of SIFT and HoG can be re-interpreted as feature maps to fit our regression network. Our network is also a part of the CNN classification model, where the former convolution blocks perform similar feature detection tasks.

The regression model can also be applied on non-image applications. The convolution operation is local, and shares similar structure with local regression models, such as the famous non-linear kernel regression. Thus many regression works can be reviewed by the CNN regression model.

**Limitations**. A limitation of our framework is, the network requires a very large training set to perform well; this is the same as other CNN models. For example, the Deep Gaze I [36] uses ImageNet pre-trained models on saliency prediction, and achieves best performance. This amounts to using a large training set in this application. Nevertheless, getting the ground truth feature maps is not as easy as getting classification labels, which aggravates the difficulty in generating training sets.

**Prevent Over-fitting**. Over-fitting is a very common problem in regression. Adding regularization terms on model parameters will prevent over-fitting and achieve better generalization ability on testing sets. In our CNN regression model, a simple $L_2$ penalty on network is applied as regularization terms. Other types of penalties, such as the $L_1$ norm, can offer some desired properties like sparsity [46, 47, 48, 49]. However, network with these penalties can be more complex to optimize. Adding regularization terms alleviates over-fitting by reducing the effective degrees of freedom (DoF) of the network [18]. For example, a large network will over-fit on simple tasks like our face detection and segmentation problem; adding penalties can reduce the effective parameters and make the network behave like a small-scale network. Determining the proper DoF or network scale for different applications is very difficult, and often based on test-and-trial. The drop-out technique [50] is a famous way to prevent over-fitting in neural networks; it is proved to be an adaptive $L_2$ regularization [51] on networks, which alleviates the test-and-trial process on penalty selection.

Another way to prevent over-fitting is the data approach. Providing huge training data to the network can also be considered as a regularization process: the training data itself covers the whole space; a network trained on the whole space data will have better generalization ability than those trained on a (small) subspace, since the test example will always lie in the space expanded by training samples. In other words, the testing phase will always be an interpolation process rather than extrapolation process; the latter is known to be much more difficult than the former. The ImageNet pre-trained model is the big data approach, and is proved to have good generalization ability in many other classification tasks. Acquiring large data is sometimes not possible; data augmentation is often applied [11] to alleviate the requirement for training data. Typical data augmentations are geometric transforms on the image dataset. This offers enlarged space (though still limited) spanned by the training images.

**Relationship with the CNN classification model**. The convolution blocks in our network are identical to those in the CNN classification model. The linear combination on convolution channels can be considered as a partially connected layer, compared to the fully connected layer in the classification model. Thus by adding another level of connection over the output feature map, we yield the CNN classification model.

The feature maps often correspond to the interested objects, which are also crucial to the classification task. We can consider the CNN classification model as an implicit feature map generating process using image class labels, whereas the regression model uses explicit feature maps generated by other techniques. The convoluted features before fully connected layers can be considered as implicit feature maps corresponding to objects. It might explain why the ImageNet pre-trained classification model is also successful in many non-classification applications.

## 6. Conclusion

We propose a CNN-based regression model trained on continuous feature maps. The regression network does not require fully connection layer, and is insensitive to input sizes. We introduce an up-sampling technique for size compatibility, and generate output feature maps through a linear combination on convolution channels. We apply this framework to face detection & segmentation and saliency prediction, and demonstrate its generalization ability in these tasks. Also, this general framework is highly related to the classification model and has the potential to be applied on variety of applications.

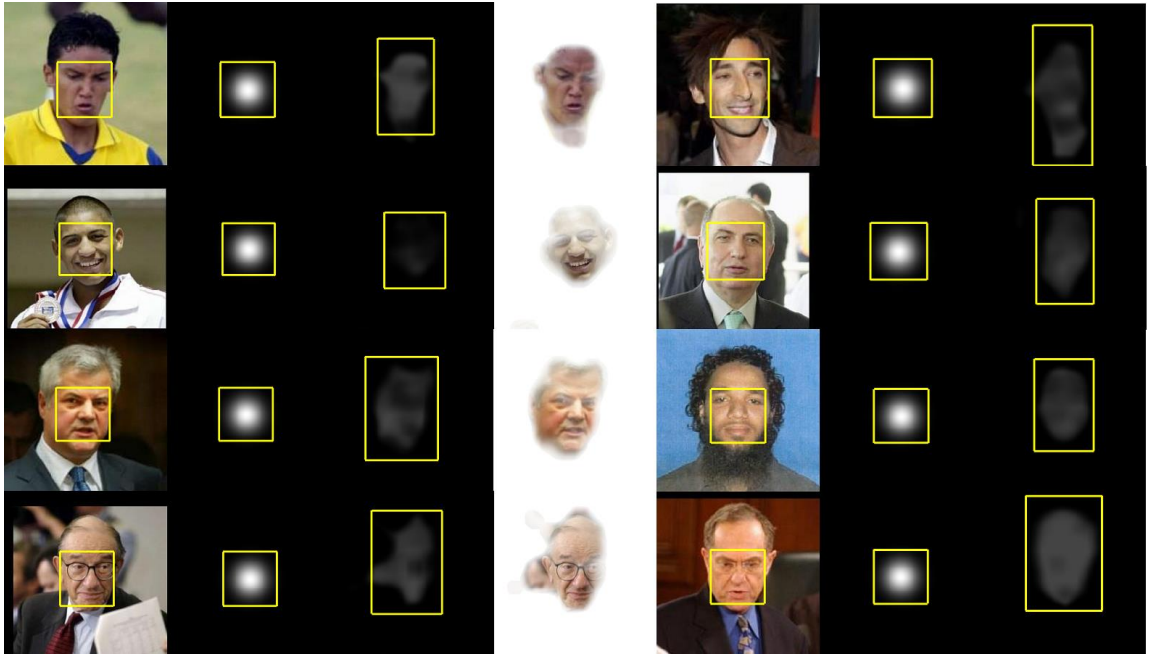

Figure 7: Additional experiment results on simultaneous face detection & segmentation

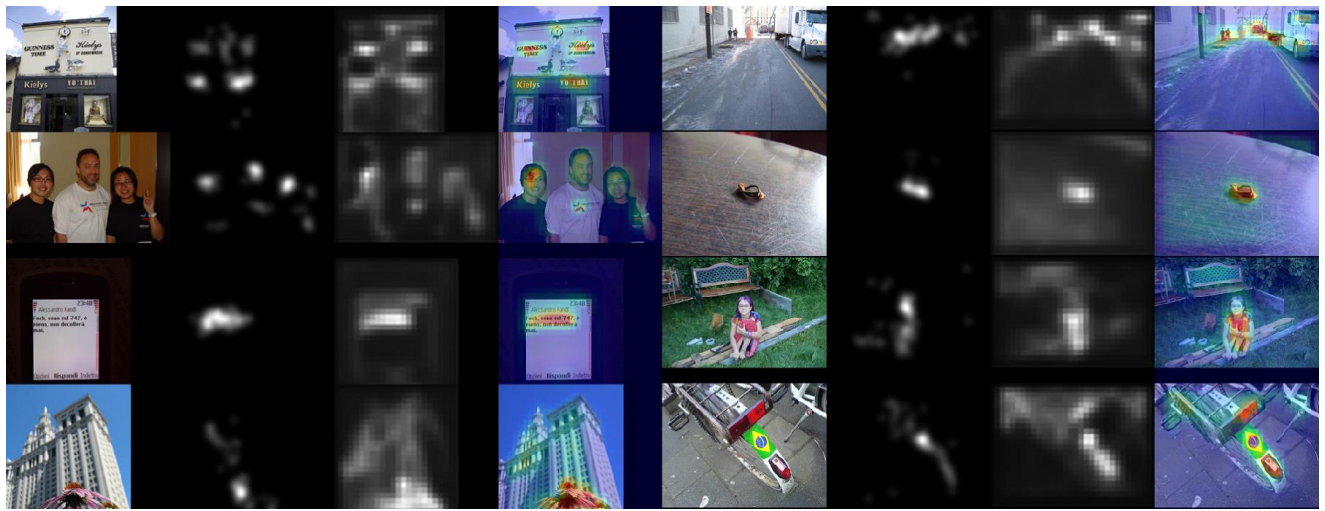

Figure 8: Additional experiment results on scene saliency prediction